\documentclass[letterpaper, 10 pt, conference]{ieeeconf}  

\IEEEoverridecommandlockouts                              

\usepackage{mathrsfs}
\usepackage{amsmath}
\usepackage{amssymb}
\usepackage{algorithm}
\usepackage[noend]{algpseudocode}
\usepackage{color}
\usepackage{cite}
\usepackage{subfigure}
\usepackage{bbm}
\usepackage{url}
\usepackage{caption}
\usepackage{gensymb}
\usepackage{makecell}
\usepackage{multirow}

\usepackage{algorithmicx}
\usepackage{algpseudocode}

\algdef{SE}[DOWHILE]{Do}{doWhile}{\algorithmicdo}[1]{\algorithmicwhile\ #1}%

\definecolor{orange}{rgb}{1,0.5,0}

\newcommand{\vtxt}[1]{}
\newcommand{\dtxt}[1]{}
\newcommand{\stxt}[1]{}
\newcommand{\etxt}[1]{}

\usepackage{xcolor}
\usepackage[normalem]{ulem}

\newcommand{\method}{Active-HOF}

\usepackage{graphicx}

\title{\LARGE \bf
Higher Order Function Networks for View Planning \\and Multi-View Reconstruction
}

\author{Selim Engin, Eric Mitchell, Daewon Lee, Volkan Isler, Daniel D. Lee
\thanks{All authors are with the Samsung AI Center - New York. Corresponding author is Selim Engin. Email: kazimselimengin@gmail.com}
}

\begin{document}

\maketitle
\thispagestyle{empty}
\pagestyle{empty}

\begin{abstract}
We consider the problem of planning views for a robot to acquire images of an object for visual inspection and reconstruction. In contrast to offline methods which require a 3D model of the object as input or online methods which rely on only local measurements, our method uses a neural network which encodes shape information for a large number of objects. We build on recent deep learning methods capable of generating a complete 3D reconstruction of an object from a single image. Specifically, in this work, we extend a recent method which uses Higher Order Functions (HOF) to represent the shape of the object. We present a new generalization of this method to incorporate multiple images as input and establish a connection between visibility and reconstruction quality. This relationship forms the foundation of our view planning method where we compute viewpoints to visually cover the output of the multi-view HOF network with as few images as possible. Experiments indicate that our method provides a good compromise between online and offline methods: Similar to online methods, our method does not require the true object model as input. In terms of number of views, it is much more efficient. In most cases, its performance is comparable to the optimal offline case even on object classes the network has not been trained on.
\end{abstract}

\section{Introduction}

\begin{figure}[ht!]
    \centering
\includegraphics[width=0.75\columnwidth]{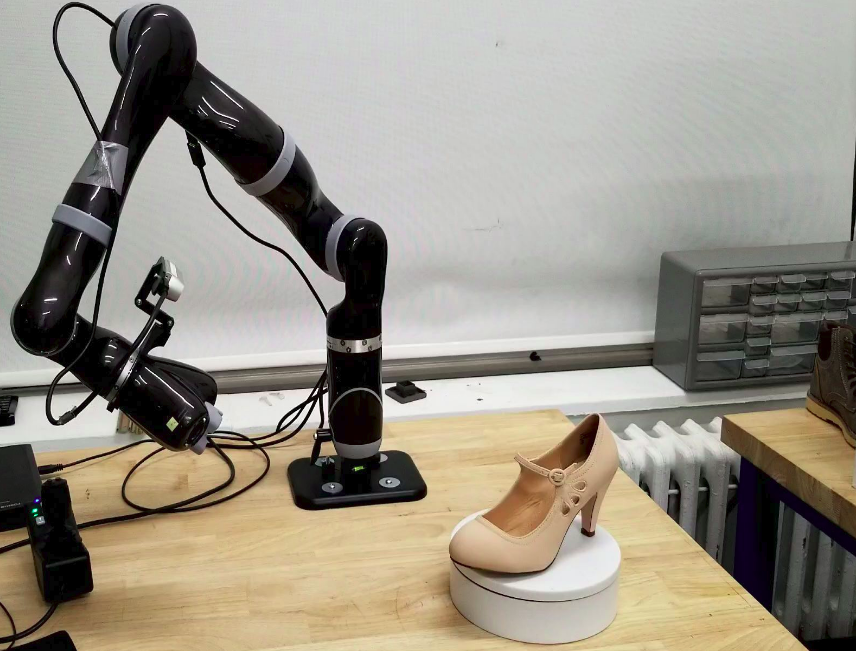}
    \caption{Our setup consists of a camera attached to the wrist of a manipulator arm and an object lying within the workspace of the robotic arm and visible from different pose configurations of the camera.}
    \label{fig:setup}
\end{figure}

The problem of choosing camera poses so as to obtain ``good" views of an object arises in numerous applications such as object inspection and 3D model capture. In this paper, we consider a scenario where a camera is mounted on a robotic actuator and tasked with acquiring views of a target object in order to reconstruct its 3D model (Figure~\ref{fig:setup}).

Approaches to view planning can be categorized into two groups based on whether the object shape is known in advance or not. In applications such as industrial part inspection, a model (and the pose) of the object is available in advance. In this case, it is possible to choose the views in an \emph{offline} fashion using the prior model. Offline view planning algorithms can provide global performance guarantees for example, in terms of the total number of views to visually cover the object. If the object is unknown, view planning must be performed in an \emph{online} fashion. The online problem is typically solved greedily where the algorithm tries to find the Next Best View (NBV) using information available so far. Since they do not require the object shape as input, online approaches are more generally applicable. However, they usually do not provide any global guarantees.

In this paper, we present a new approach which combines the strengths of online and offline approaches. Our approach builds on recent progress on single-view reconstruction methods in which prior information about shapes of objects is encoded by a neural network capable of generating a 3D model of the object from a single image~\cite{fan2017point, tatarchenko2017octree, groueix2018atlasnet, mees2019self}. 

State of the art methods are successful in reconstructing a large number of objects from a fairly large number of classes.
However, they also have some disadvantages:
They are usually evaluated on synthetic datasets, and their generalization to real world data is often not very successful.
Furthermore, they  generate coarse reconstructions.
For instance,~\cite{mees2019self} outputs a $32^3$ voxel grid to represent the objects. Also, most other methods are limited to the class of objects they are trained on.



We show that a similar approach can be used to encode prior information which can then be used for planning views effectively. These images afterwards can be used to generate fine scale reconstructions using traditional approaches or other visual inspection tasks such as quality inspection. 

\begin{figure}[ht!]
  \centering
  \includegraphics[width=\columnwidth]{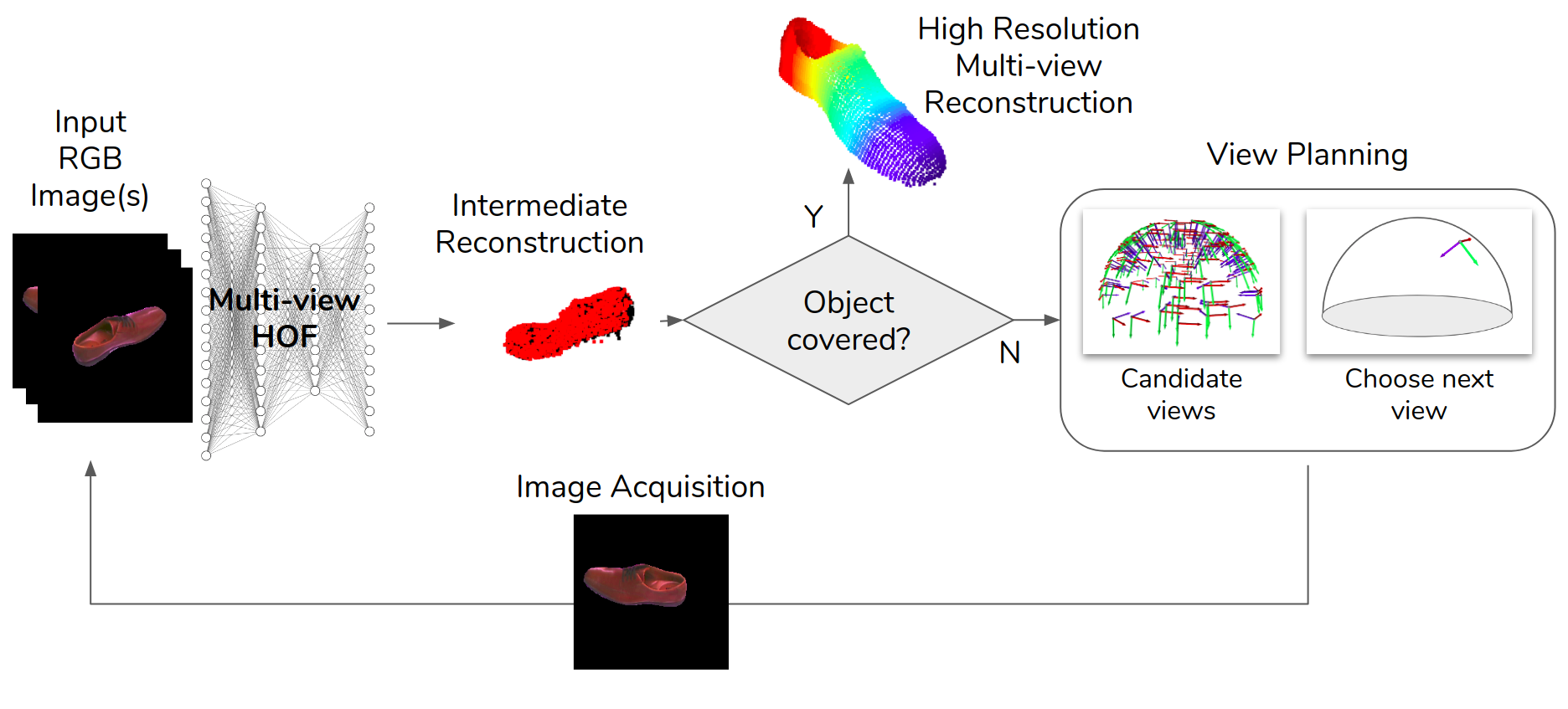}
  \caption{\textbf{Overview of our method.} Starting with a masked RGB image of the scene, we perform 3D reconstruction to obtain a global intermediate geometry of the object. This shape is used for guiding the view planning algorithm and the images taken from the visited viewpoints are incorporated for refining the reconstruction, until the object surface is covered. }
    \label{fig:overview}
\end{figure}

Our method, which we call \textbf{\method{}},  starts with an RGB image and performs single-view 3D reconstruction to obtain an intermediate geometry of the object which is used for selecting the following views (Figure~\ref{fig:overview}).
With each additional view, the intermediate reconstruction is refined by incorporating all previous images using our novel multi-view reconstruction network presented in Section~\ref{sec:multi-hof}.  This process is repeated until the visible surface of the object is covered.



Our approach is more general than classical offline methods since once trained, it does not require an object model as input. We also show that it is more effective than online approaches in terms of the number of views. 
To summarize, in this paper we:
\begin{itemize}
    \item present a method to 3D reconstruct an object (including parts unseen from any view) by incorporating multiple measurements and prior information in a seamless fashion
    \item establish a connection between visibility and reconstruction quality of the output produced by this  method. This relationship is important because at the moment, there is no established method to quantify the uncertainty of the output of deep learning based reconstruction algorithms 
    \item show that view planning for covering the surface of an object using \method{} is more efficient in terms of number of images compared to an information maximization-based online approach.
\end{itemize}

\section{Related Work}

We divide the related work into two groups. We start with object inspection and then present relevant work on object reconstruction.

\subsection{Object Inspection}

In the object inspection problem the goal is to cover the visible surface of the object using the smallest possible number of views. Typical motivations for this problem include identifying building cracks, detecting diseases in plants, and painting cars. An overview of view planning methods can be found in the survey~\cite{scott2003view} and more recently in~\cite{bajcsy2018revisiting}.

Earlier works have concentrated on a model-based approach, where the model of the object to be inspected is given beforehand~\cite{scott2003view, tarbox1995planning}.
Jing et al. uses a sampling-based approach for generating viewpoints and builds a potential field to decide the viewing directions~\cite{jing2016sampling}.
Inspecting complex 3D environments studied in~\cite{englot2012sampling} uses an RRT-like algorithm, which is demonstrated in a field experiment to inspect ship hulls.
Methods in this line of work performs view planning in an offline manner using the given 3D model of the scene.

In the absence of a prior 3D shape knowledge, view planning needs to be performed in an online fashion
as information becomes available.
As one of the earliest studies dealing with this problem, Maver and Bajcsy proposed a method using occlusions and their contours in the scene to choose the next views~\cite{maver1993occlusions}.
Banta et al.~\cite{banta1995best} presented an NBV algorithm that locates the candidate viewpoints on rays emanating from high curvature points using a range sensor.
Deciding where the agent should go to explore the environment in Simultaneous Localization and Mapping (SLAM) settings is also studied as an NBV problem in numerous works~\cite{gonzalez2002navigation, grabowski2003autonomous}.
Several recent studies also proposed methods for the task of exploration using an information theoretic approach~\cite{delmerico2018comparison, julian2014mutual, jadidi2015mutual}.

\subsection{Object Reconstruction}

The 3D reconstruction of objects is a well-studied problem in the field of robotics.
With the availability of commercial depth cameras, state of the art methods such as KinectFusion~\cite{newcombe2011kinectfusion} for rigid, and DynamicFusion~\cite{newcombe2015dynamicfusion} for non-rigid scenes have achieved great success in generating dense object reconstructions.
These methods rely on depth readings and they may need many views from the scene in order to reconstruct the scene effectively.

Recently, predicting the 3D shape of the object using a single or a few views has received a significant attention.
As one of the earlier works using a learning-based approach for this problem,~\cite{choy20163d} presents a recurrent neural network architecture that encodes the input images directly onto the voxels of an occupancy grid. 
Other types of object representations, such as point sets~\cite{fan2017point}, octrees~\cite{tatarchenko2017octree}, and meshes~\cite{wang2018pixel2mesh} have also been studied in the context of object reconstruction.


\section{Problem Formulation}

Our setup consists of a camera $C$  mounted on a manipulator arm. The system is calibrated so that the coordinate frame of the camera relative to the base frame is known. The scene is static and contains a single object. For convenience, we use voxel grids to represent 3D reconstructions.

The visibility of a voxel $x_i \in \mathcal{X}$ in the camera frame from a viewpoint $V$ is given by a boolean function $vis(x_i, V)$, whose value is 1 if the point is directly visible from $V$, and 0 otherwise.
The set of voxels visible from a viewpoint $V$ is denoted as $\mathcal{X}_V \subseteq \mathcal{X}$, with $|\mathcal{X}_V| = \sum_{i=1}^n vis(x_i, V)$.

The viewing space $\mathcal{W}$ (the set of candidate camera centers) is a hemisphere centered near the object's origin. The camera pose is chosen in such way that the optical axis ($Z$) points toward the center. 
The object is assumed to be  \textit{externally-visible} from $\mathcal{W}$: for each point $p$ on the object surface there exists a camera viewpoint $V \in \mathcal{W}$ such that $vis(p, V) = 1$. 
If this is not the case, our algorithm can cover only the visible points.




Our view planning methodology requires a reconstruction algorithm to generate intermediate 3D models as images become available. In this paper, we generalize the method presented in~\cite{mitchell2019higher} to multi-view inputs (Section~\ref{sec:multi-hof}) and use it to generate 3D models which are also represented as voxel grids. We denote the measurement for the voxel $x_i$ generated using only the $k$-th frame by $z_i^k$. Similarly, $Z_i^{1:k} = \{z_i^1, \dots, z_i^k \}$ is the measurement using the first $k$ measurements for $x_i$. Each measurement for a voxel $x_i$ is a random variable whose realization is $z_i^k$, at all frames $k$.







We are now ready to state our problem. 

\textbf{Problem Statement:}
Given a masked RGB image of an externally-visible object $O$ captured by a movable camera and a desired percentage of the surface to be covered, find the smallest set of camera viewpoints such that each point $p$ on the surface of $O$ is visible from at least one viewpoint. 





\section{Method}

In this section, we start with an overview of our method followed by a description of the network architecture we use for multi-view reconstruction. Finally, we present the dataset we use to train our networks.

\subsection{Overview}

Our method takes as input a masked RGB image $I$ and a desired percentage for the coverage of the object surface.
The main loop of \method{} is given in Algorithm~\ref{algo:main}. The algorithm works as follows.
First, a point set prediction $P$ is generated by feeding $\mathcal{I} = \{I\}$ to the \emph{Multi-HOF} network.
Then, the set of visible points $P_V \subseteq P$ from the current viewpoint $V$ is computed using a ray-shooting-like method based on~\cite{katz2007direct}.

The algorithm then proceeds with voxelizing the point set $P$, and labeling each voxel as visible or not depending on the visibility of the point corresponding to that voxel. If there are more than one point falling into a voxel, we take the label of the larger set of points.

Suppose $\mathcal{X}$ and $\mathcal{X}_V$ are the voxelization of the entire point set $P$ and the visible set $P_V$, respectively, and let $\mathcal{X}_V^{1:k}$ be the union of visible voxels from all past viewpoints $V_{1:k}$. \etxt{The sentence is sort of dangling- which paragraph is it supposed to go with?}
\method{} continues by sampling a set of $n$ viewpoints $\mathcal{V}$ from the viewing space $\mathcal{W}$, and chooses the viewpoint $V^*$ maximizing the number of previously invisible set of voxels as the next view, 
$V^* = \arg \max_{V' \in \mathcal{V}} {|\mathcal{X}_{V'} \cap (\mathcal{X} - \mathcal{X}_V^{1:k})|}$.


A principled way of determining $n$, the number of sampled viewpoints, is through the notion of $\varepsilon$-nets~\cite{haussler1987epsilon, isler2004sampling}.
Let $V(x)$ denote the region on the viewing space $\mathcal{W}$, where the point $x$ is directly visible.
If there is an $\varepsilon$ such that $V(x)$ contains a disk of area $\varepsilon$ for all $x \in \mathcal{X}$, then from~\cite{haussler1987epsilon} we know that $n = O(\frac{A}{\varepsilon} \log \frac{A}{\varepsilon})$ samples suffice to cover all the points, where $A$ is the surface area of the viewing space.

\begin{algorithm}
\caption{\textsc{\method{}}}
\begin{algorithmic}[1]
  \Require {$I_1$: masked RGB image of the object; $\alpha$: desired coverage percentage; $n$: number of sampled viewpoints}
  \Ensure {$P$: a high resolution 3D reconstruction; $\mathcal{I}$: a set of images covering $\alpha$-portion of the object surface}
  \State $k \leftarrow 1$; $\mathcal{I} \leftarrow \{I_1\}$; $\mathcal{X}_V^{1:k} \leftarrow \varnothing$
  \State $V \leftarrow $ Current viewpoint
  \Do
    \State $P \leftarrow $ \emph{Multi-HOF}($\mathcal{I}$)
    \State $P_V \leftarrow$ Visible points of $P$ from viewpoint $V$
    \State $\mathcal{X}, \mathcal{X_V} \leftarrow $ Voxelize $P$ and $P_V$, respectively
    \State $\mathcal{X}_V^{1:k} \leftarrow \mathcal{X}_V^{1:k} \cup \mathcal{X_V}$
    \State Sample a set of $n$ viewpoints $\mathcal{V}$ from $\mathcal{W}$ 
    \State $V^* \leftarrow \arg \max_{V' \in \mathcal{V}} {|\mathcal{X}_{V'} \cap (\mathcal{X} - \mathcal{X}_V^{1:k})|}$
    \State $V \leftarrow V^*$; $k \leftarrow k + 1$; $\mathcal{I} \leftarrow \mathcal{I} \cup \{I_k\}$
  \doWhile{$|\mathcal{X}_V^{1:k}| \leq \alpha \cdot |\mathcal{X}|$} 
\end{algorithmic}
\label{algo:main}
\end{algorithm}

\subsection{Multi-HOF: Learning Multi-view 3D Reconstruction}
\label{sec:multi-hof}
Applying neural networks, particularly convolutional neural networks, to the task of 3D reconstruction is an active area of research. While convolutional networks have proven extremely effective in 2D image domains, extending them to 3D dataset is non-trivial. In this work, we build on the Higher-Order Functions (HOF) architecture for point set-based 3D reconstruction \cite{mitchell2019higher}. HOF defines a parameterized function $f_\theta:\mathbb{R}^3 \rightarrow \mathbb{R}^3$ with parameters $\theta$ and a function $g:\mathbb{R}^{3\times d \times d}\rightarrow \mathbb{R}^{|\theta|}$. Thus $g$ is a `higher-order` function that maps a $d\times d$ RGB image to the parameters $\theta$ of the `lower-order' function $f$, which defines a transformation of points within the unit sphere to the surface of an object. Our reconstruction is defined as $\{f_\theta(\mathbf{x}_i): \mathbf{x}_i \in X\}$, where $X$ is a set of points randomly and uniformly sampled from the interior of the unit sphere.

\begin{figure}[ht!]
    \centering
\includegraphics[width=0.9\columnwidth]{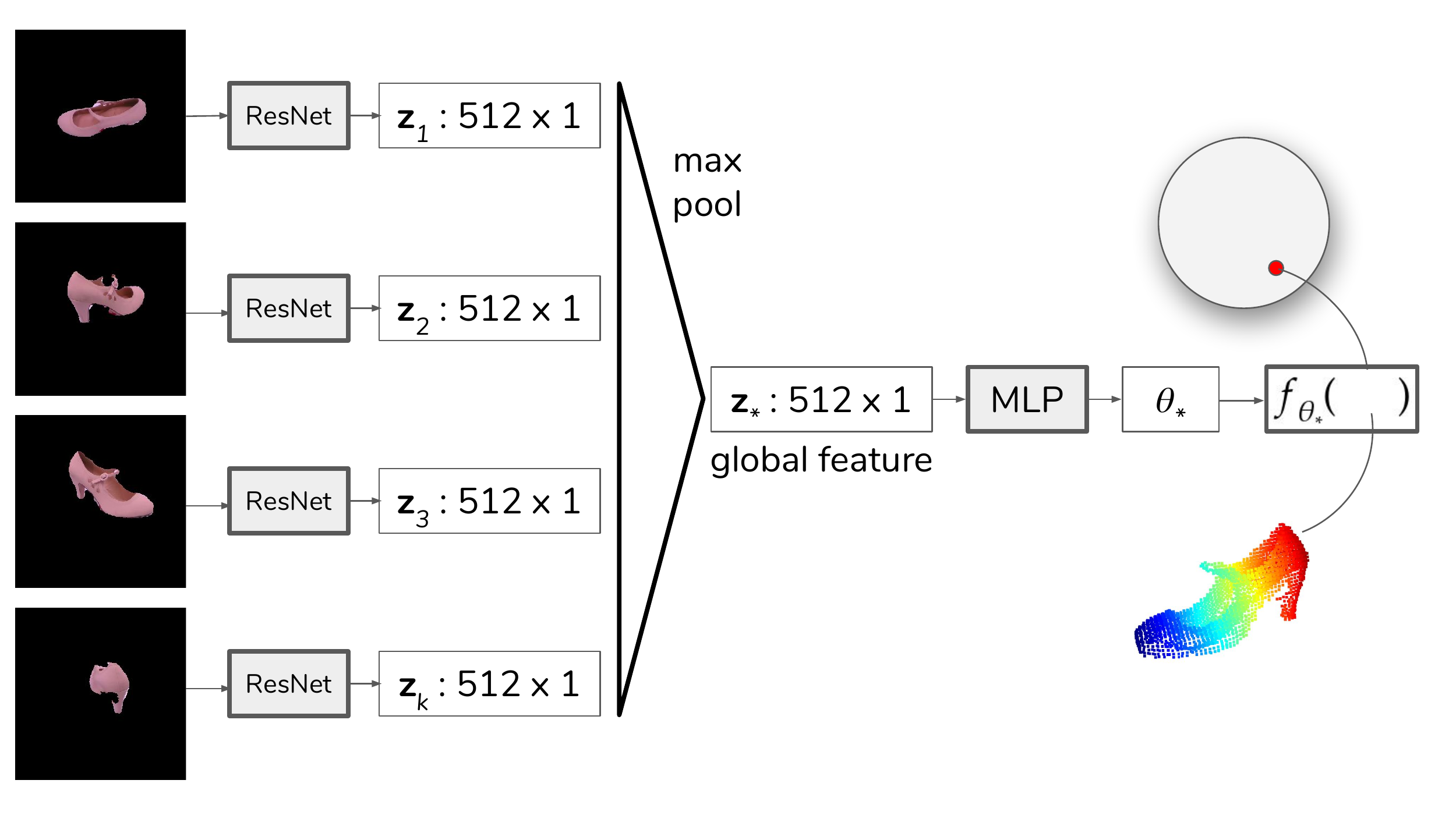}
    \caption{\textbf{Multi-view HOF network architecture.} Each captured image is fed into a ResNet-34~\cite{he2016deep} network whose output is a 512 dimensional encoding vector. After pooling the encodings, Multi-HOF produces a function $f_{\theta_*}$ which is used to map a point from a unit sphere onto the surface of the object.
    \label{fig:architecture}
    }
\end{figure}

To extend HOF to multiple views, we decompose the function $g$ into two functions $g_1:\mathbb{R}^{3\times d \times d}\rightarrow \mathbb{R}^c$ and $g_2:\mathbb{R}^c \rightarrow \mathbb{R}^{|\theta|}$. $g_1$ is an image encoder that produces a fixed-length encoding $\mathbf{z}_i$ of dimension $c$ for each input image $I_i$. See Figure~\ref{fig:architecture}. After computing each $\mathbf{z}_i$, we compute a `global' encoding vector, also of dimension $c$, where the value at each index is the largest activation at that index among each of the image encodings. The pooling step yields a single `global' encoding $\mathbf{z_*} \in \mathbb{R}^c$ representing the entire set of images. From $\mathbf{z_*}$, we compute $\theta_* = g_2(\mathbf{z_*})$ and the final object reconstruction using $f_{\theta_*}$ as in the original HOF formulation.

\subsection{Dataset}


For training and evaluating our methods we need a dataset of objects with real world data.
To that end, we have a dataset consisting of RGB images, segmentations, pointclouds and camera to object relative poses of real shoes and mugs.
The RGB images are taken from a hemisphere centered at the origin of the object, and the camera viewpoint optical axes point toward the object origin as well.

Our dataset is one of the few real object datasets for the purpose of 3D reconstruction, which usually requires a large number of objects within the same category.
We have 40 shoes in our shoe dataset, of which 25 is in the training and 15 is in the test set.
The number of models in our mug dataset are 6 for the training set and 5 for the test set. 
A more detailed description of an earlier version of our dataset can be found in~\cite{tosun2019pixels}.
For this paper, we expanded our dataset to include a novel class of object (mugs) and we report results on generalization of our method to unseen categories in Section~\ref{sec:exp_novel_class}.

\section{Analysis}

In this section, we analyze our method through a series of experiments. 


\subsection{Experiment 1: Relationship between visibility and reconstruction quality}

The uncertainty in geometric reconstruction methods such as triangulation can be established geometrically using methods such as GDOP~\cite{kelly2003precision}. Unfortunately, at present there are no such metrics for deep-learning based reconstruction methods. Therefore, in our first experiment, we investigate whether visibility can be used as a proxy for reconstruction uncertainty.  To do this, we quantify the impact of visibility in single-view 3D reconstructions.
Specifically, \emph{we investigate whether the uncertainty (entropy) is lower for the visible points compared to the entire shape.}
We take a small subset of possible viewpoints, perform single-view reconstruction from each view, and compare the cumulative entropy scores of the entire object against the voxels visible from all selected views.

\begin{figure}[ht!]
\centering     
\subfigure[Views chosen from the back of the shoe used for reconstruction
\label{fig:exp1-img}]{\includegraphics[width=0.95\columnwidth]{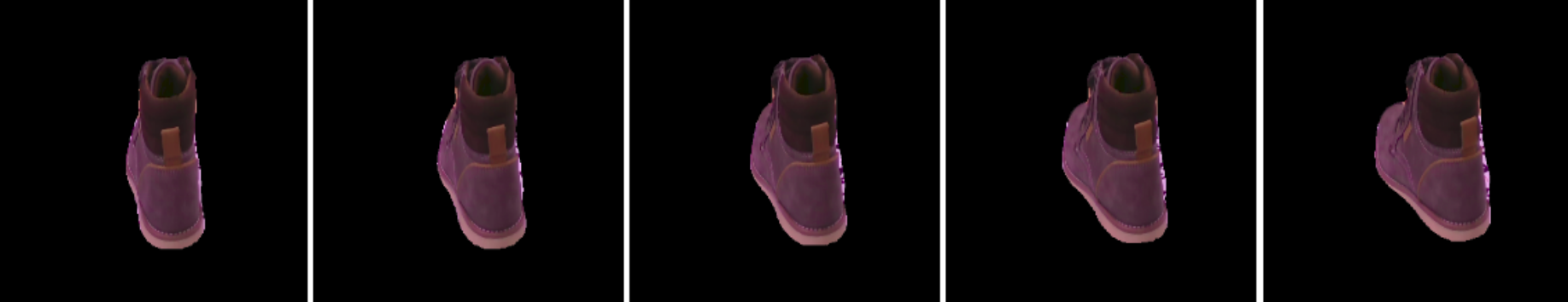}}
\subfigure[Entropy over the entire reconstruction \label{fig:exp1-entire}]{\includegraphics[width=0.45\columnwidth]{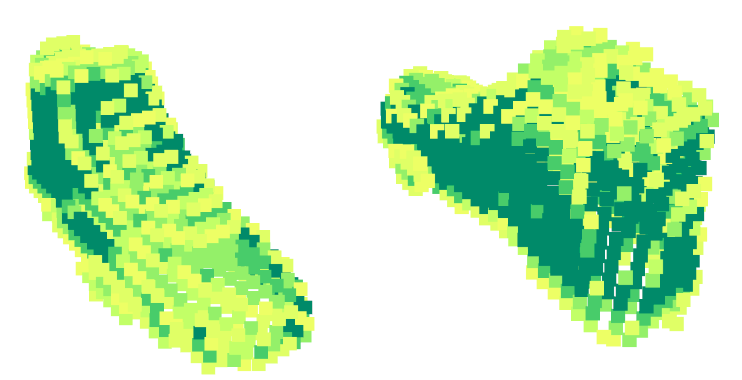}}
\subfigure[Entropy over the visible parts of the reconstruction \label{fig:exp1-visible}]{\includegraphics[width=0.45\columnwidth]{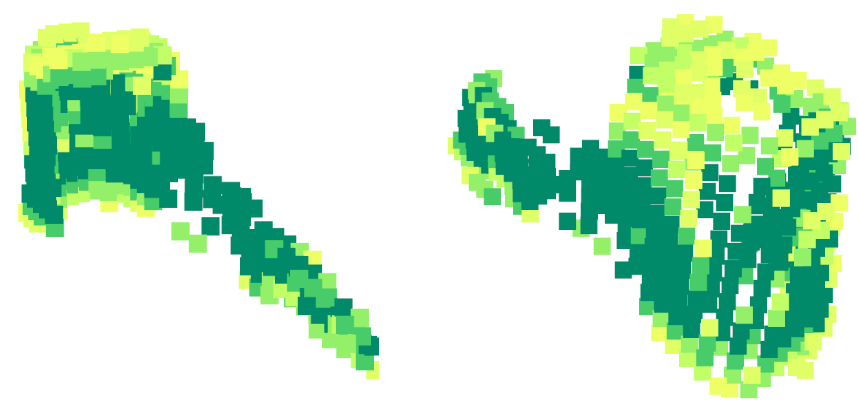}}
\caption{ 
Two views of the voxelized entropy maps over the entire and visible parts of the reconstruction.
The entropy of the visible part is lower compared to that of all the voxels.
Darker to brighter colors indicate an increase in the entropy. }
\label{fig:exp1}
\end{figure}

Specifically, we select a small number of consecutively ordered viewpoints from the dataset (about $5\degree$ between each pair) and reconstruct the object using RGB images from each viewpoint (see Figure~\ref{fig:exp1-img}).
We represented the predicted object either as $40^3$ or $80^3$ voxels of edge length $10$mm and $5$mm, respectively.
Using $k = 5, 8, 11$ viewpoints, Table~\ref{tab:exp1-entropy} shows the entropy scores of the reconstruction's all (visible + invisible) and visible-only voxels.


\begin{table}[ht!]
    \centering
     \begin{tabular}{| c | c | c | c |}
    \hline
    Dim. & \# views & Entire Obj. Entr. & Visible Obj. Entr. \\ \hline
    40 & 5 & 0.0118 / 0.09 & \textbf{0.0069 / 0.06} \\ \hline
    40 & 8 & 0.0133 / 0.10 & \textbf{0.0084 / 0.07} \\ \hline
    40 & 11 & 0.0135 / 0.09 & \textbf{0.0090 / 0.07} \\ \hline
    \hline
    80 & 5 & 0.0055 / 0.05 & \textbf{0.0033 / 0.04} \\ \hline
    80 & 8 & 0.0061 / 0.06 & \textbf{0.0036 / 0.04} \\ \hline
    80 & 11 & 0.0066 / 0.06 & \textbf{0.0038 / 0.03} \\ \hline
    \end{tabular}
    \caption{Mean and standard deviation comparison of voxel entropies in the entire object vs. visible parts of the object}
\label{tab:exp1-entropy}
\end{table}

After performing 3D reconstruction from each viewpoint, we voxelize the predicted pointclouds and compute a probability measure using a simple voting scheme.
The number of viewpoints voting for the occupancy of each voxel divided by the number of measurements gives us a proxy for the probability of the cell being full.


The entropy map of the reconstruction is shown in Figure~\ref{fig:exp1-entire} and~\ref{fig:exp1-visible}, where darker colors indicate lower entropy scores.

We find that the uncertainty of the visible parts of the reconstruction is lower compared to the entire reconstruction.
This suggests that the reconstruction performance is better for the visible points.


\subsection{Experiment 2: Visibility-based Coverage}

After showing the importance of the visibility for reconstruction, the next question we investigate is: \textit{How valuable is a global reconstruction as a prior for view planning?}
In this experiment, we compare our method against an information maximization-based approach.

If there is no prior knowledge about the scene, exploration of the environment using an information theoretic formulation is a common approach in the literature~\cite{delmerico2018comparison, julian2014mutual, jadidi2015mutual}.
These methods typically rely on local measurements to update their priors on the scene using a pre-defined sensor measurement model.
The posterior map is then used for choosing the next sensor location and optimizing an objective function such as conditional entropy or mutual information.

The information-driven approach we use as a baseline in this experiment operates on the scene voxelization $\mathcal{X}$, with each voxel probability initialized as $p(x_i) = 0.5$ meaning no prior knowledge.
We apply Bayesian filtering and use log-odds update~\cite{thrun2005probabilistic} to compute the posterior probabilities of each voxel, assuming independence between the random variables $x_i \in \mathcal{X}$. 


\begin{table}[ht!]
    \centering
     \begin{tabular}{| c | c || c | c |}
    \hline
    Coverage & \thead{Vis. max. \\(GT points)} & \thead{Info. max. \\(Depth GT points)} & \thead{\method{} \\(Pred. points)} \\ \hline
    70\% & \textbf{2.44 / 0.53} & 4.72 / 3.77 & \textbf{2.65 / 0.67} \\ \hline
    80\% & \textbf{3.18 / 0.83} & 8.92 / 7.05 & 5.67 / 3.26 \\ \hline
    90\% & \textbf{5.11 / 2.64} & 16.27 / 6.23 & 6.33 / 2.95 \\ \hline
    \end{tabular}
    \caption{Average number of views used to cover a desired percentage of the object surface using visibility- and information-based methods} 
    \label{tab:coverage}
\end{table}

We compare \method{} against two baselines with respect to the number of views to cover a desired percentage of the object's visible surface. 
Our first baseline is the same with \method{} except that it performs visibility maximization on the ground truth (GT) point set, instead of predicting it.
The second baseline is an information maximization method that uses the true depth readings from the object.

Setting the desired percentage to be 70, 80 and 90\%, the number of viewpoints necessitated by each approach is shown in Table~\ref{tab:coverage}.
We note that the local information-driven method was sometimes unable to cover the desired percentage of the object surface points, especially for intricately shaped objects like high-heels. In our experiments we used a hard limit of 20 views after which point the execution terminates.

\begin{figure}[ht!]
    \centering
\includegraphics[width=0.75\columnwidth]{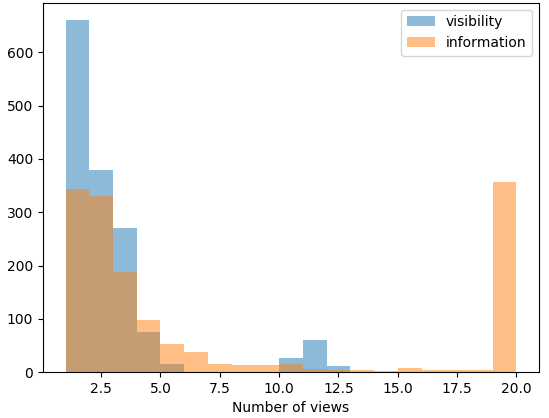}
    \caption{The number of viewpoints used by our visibility-based method and a local information-driven approach}
    \label{fig:hist_num_views}
\end{figure}

\begin{figure}[ht!]
    \centering
    \includegraphics[width=0.85\columnwidth]{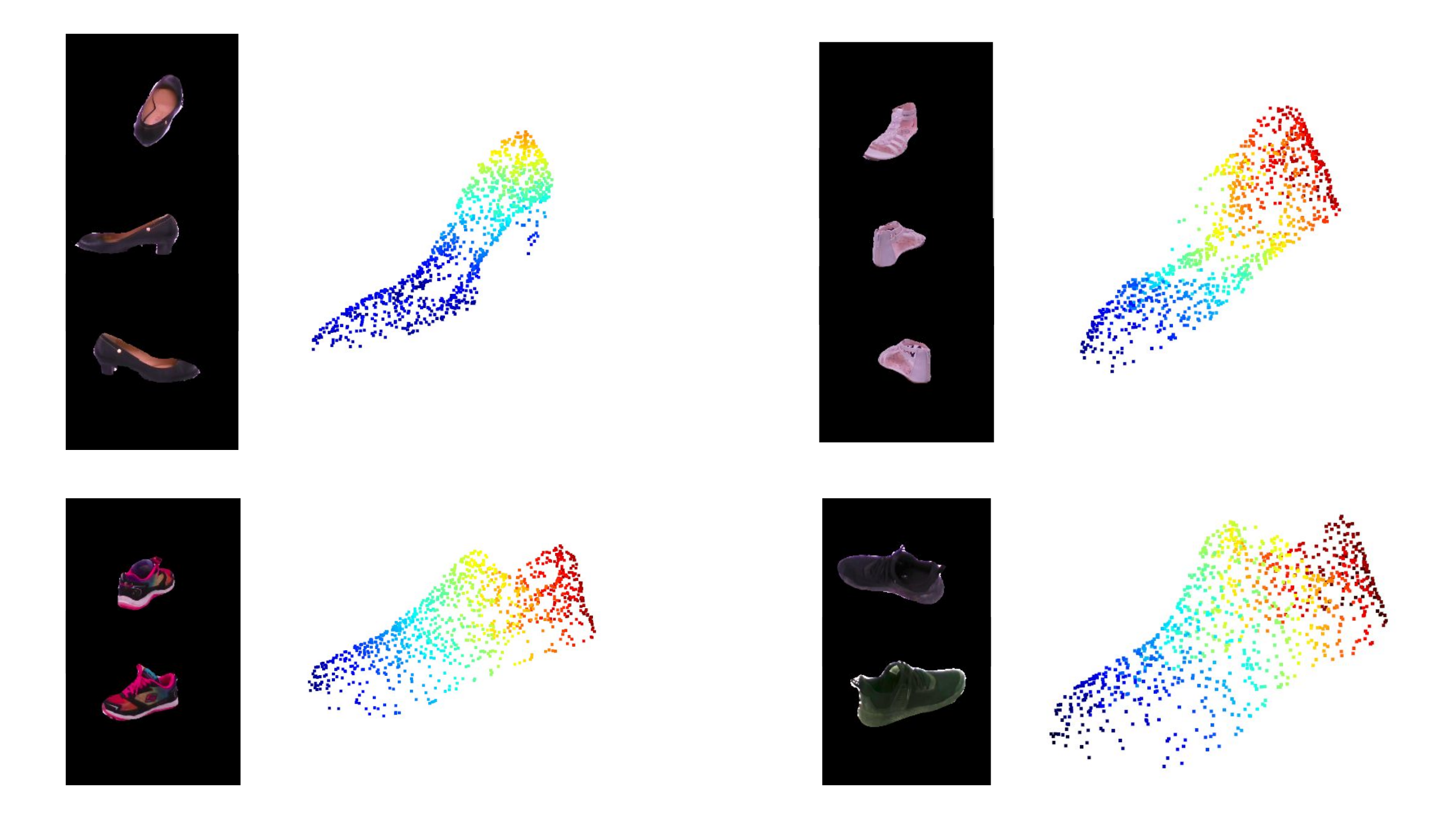}
    \caption{Acquired images to cover 80\% of the shoe surface and the resulting high resolution 3D reconstructions}
    \label{fig:images_recons}
\end{figure}

The histogram in Figure~\ref{fig:hist_num_views} shows the counts for the number of views using the information-driven baseline and our approach.
The images taken from the selected viewpoints using \method{} for some of the shoes in our test set is shown in Figure~\ref{fig:images_recons}.
Our results indicate that using a global 3D shape prediction to guide the view planner performs favorably compared to online methods that operate on local information.
Furthermore, we find that the performance of \method{} is comparable to the visibility maximization baseline that uses the true points, for low coverage rates in particular.

\subsection{Experiment 3: Single vs. Multi-view Reconstruction}

In this experiment we first evaluate the contribution of additional views on the prediction quality of learned multi-view reconstructions. Then, we identify if it is worth refining the intermediate geometry for view planning.

A common metric for measuring the quality of 3D reconstructions is Intersection over Union (IoU), which we use to compare the similarity between the true and predicted voxel grids.
Table~\ref{tab:k_hof} shows the average IoU results for our HOF network trained using up to and including 4 images. We see an improvement in the IoU performance of our method with each additional image. This demonstrates the value of extra images for the task of 3D object reconstruction.

\begin{table}[ht!]
    \centering
     \begin{tabular}{| c | c | c | c |}
    \hline
    k = 1 & k = 2 & k = 3 & k = 4 \\ \hline
    0.429 / 0.12 & 0.434 / 0.12 & 0.436 / 0.12 & 0.438 / 0.11 \\ \hline
    \end{tabular}
    \caption{Average IoU results for single-view HOF (k=1) vs. multi-view HOF with k = 2, 3, 4 images}
    \label{tab:k_hof}
\end{table}

We also evaluate our network by inputting more images at a time than were used to train the network.
Multi-HOF generates an encoding vector of each image separately, and the global feature vector is computed by max pooling the individual encodings, which is an unordered set operation.
This allows the network to a variable number of images at inference time, which is useful for our method as the number of views needed is unknown.

The generalization results of Multi-HOF to higher number of images at test time is shown in Table~\ref{tab:k_hof_generalize}.
We see that the reconstruction quality can be maintained or even improved (in the case of k = 5) by using a larger number of images, beyond what the network was trained on.
However, we also see that the results start to degrade after a few additional images.
This could be due to fusing of too many encoding vectors which might lead the predictions lose their sharp geometric features.
It should also be noted that for our setting, having more images does not indefinitely improve the visibility of the object due to the notion of diminishing marginal utility: the contribution of each view for the visibility of regular geometric shapes decreases on average, given a sufficient number of sample viewpoints.


\begin{table}[ht!]
    \centering
     \begin{tabular}{|c|c|c|c|c|c|c|c|}
    \hline
    \textbf{k} & 5 & 6 & 7 & 8 & 9 & 10 & 11 \\ \hline
    Mean & 0.439 & 0.434 & 0.437 & 0.436 & 0.434 & 0.435 & 0.433 \\ \hline
    $\sigma$ & 0.12 & 0.11 & 0.12 & 0.11 & 0.12 & 0.12 & 0.11 \\ \hline
    \end{tabular}
\caption{Mean and standard deviation ($\sigma$) of the IoU results for k-view HOF generalization to higher values of $k$}
    \label{tab:k_hof_generalize}
\end{table}

Next, we analyze the significance of 
doing the refinement update for the intermediate reconstruction of the object.
Since we can perform a global 3D reconstruction using a single image, it is natural to ask if updating the reconstruction with new measurements improve the view planning performance.
We refer to the version of our method that uses only the reconstruction from the first frame and never updating it as \emph{static}, and our method as \emph{dynamic} reconstruction.
Intuitively, even a rough estimation of the object shape can be very helpful guiding the view planning method, especially for lower desired coverage percentages.
As the desired percentage increases, however, the intermediate geometry of the object becomes less reliable and the number of measurements to cover the surface of the object gets larger.

We present the number of views taken by each approach to cover various percentages of the object surface in Table~\ref{tab:static-v-dynamic}.
Note that as the desired coverage rate increases the number of images required by the static and dynamic reconstruction approaches deviate significantly.

\begin{table}[ht!]
    \centering
     \begin{tabular}{| c | c | c | c |}
    \hline
    Coverage & Static Reconstruction & Dynamic Reconstruction \\ \hline
    80\% & 2.06 / 0.242 & 2.02 / 0.156 \\ \hline
    85\% & 2.15 / 0.357 & 2.06 / 0.242 \\ \hline
    90\% & 2.60 / 0.538 & 2.35 / 0.476 \\ \hline
    95\% & 3.35 / 0.823 & \textbf{2.75 / 0.541} \\ \hline
    \end{tabular}
    \caption{Average number of views taken to cover a desired percentage of the predicted object surface using static and dynamic reconstructions for our view planning method}
    \label{tab:static-v-dynamic}
\end{table}

\subsection{Experiment 4: Generalization to Novel Classes of Objects}
\label{sec:exp_novel_class}

In this experiment we investigate how our view planning method performs on an unseen object class such as mugs.

We start by performing an IoU comparison between the performances of Multi-HOF trained on mug-only and shoe-only datasets, evaluated on the mug test set.
The results of this comparison using up to 7 images are presented in Table~\ref{tab:k_hof_mug}.

We see that the generalization of the network trained on the shoe dataset to a novel class of objects for the task of multi-view 3D reconstruction is not very well.
Nevertheless, the reconstruction quality steadily improves for both networks as the number of images is increased.

\begin{table}[ht!]
    \centering
     \begin{tabular}{| c | c | c | c | c | c | c | c |}
    \hline
\textbf{Class} & k = 1 & k = 2 & k = 3 & k = 4 & k = 5 & k = 6 & k = 7 \\ \hline
    Mug & 0.559 & 0.561 & 0.560 & 0.562 & 0.570 & 0.569 & 0.568 \\ \hline
    Shoe & 0.167 & 0.170 & 0.174 & 0.177 & 0.179 & 0.180 & 0.182 \\ \hline
    \end{tabular}
\caption{Mean IoU results for Multi-HOF trained on mugs-only and shoes-only datasets evaluated on the mug test set using up to k = 7 images}
    \label{tab:k_hof_mug}
\end{table}

We then analyze the performance of \method{} trained on shoes evaluated with the mug dataset.
In contrast to the task of reconstruction, we find that for view planning the novel class generalization performance is much more competitive. 
The results for this comparison is shown in Table~\ref{tab:dynamic_at_novel}. The results indicate that the intermediate shape reconstructions provide valuable prior information even when the actual reconstruction is not very accurate.

\begin{table}[ht!]
    \centering
     \begin{tabular}{| c | c | c | c |}
    \hline
    Coverage & Mug dataset & Shoe dataset \\ \hline
    70\% & 3.98 / 1.01 & 4.90 / 1.84 \\ \hline
    80\% & 6.18 / 2.15 & 7.36 / 2.41 \\ \hline
    90\% & 9.04 / 2.55 & 9.86 / 2.18 \\ \hline
    \end{tabular}
    \caption{Number of views for covering a desired percentage of the GT mug surface using \method{} trained on mug and shoe datasets}
\label{tab:dynamic_at_novel}
\end{table}

\subsection{Experiment 5: Evaluation on a Robot System}
The experiments reported so far were performed on data collected in an offline fashion using the set up shown in Figure~\ref{fig:setup}.
In our final set of experiments, we report results from real-time operation of the system.  We use an Intel RealSense D435 camera mounted on a Kinova Jaco2 manipulator.  
The object is placed at a known position relative to the robot base frame. The starting configuration of the camera is arranged to be pointing toward the origin of the object.

To segment the scene we perform plane subtraction using the depth images, however state of the art image segmentation methods such as~\cite{he2017mask} based on RGB images can also be used.
We evaluate our method on shoes from both training and test sets of our dataset and report the number of views taken to cover a desired percentage of the predicted object surface.
Table~\ref{tab:robot_experiment} shows the mean and standard deviation of the number of views taken to cover 80\% of the shoe surface.
We see that the performance of our method on the test set is comparable to that of the training set shoes validating our earlier results.  


\begin{table}[ht!]
    \centering
     \begin{tabular}{|c|c|c|c||c| c | c |}
    \hline
     & Heel & Sneaker & Derby & Boot & Runner & Kid\\ \hline
    Mean & 5.16 & 5.20 & 3.85 & 6.20 & 4.50 & 4.50 \\ \hline
    $\sigma$ & 1.21 & 1.93 & 0.64 & 1.46 & 1.24 & 1.81 \\ \hline
    \end{tabular}
    \caption{Mean and standard deviation ($\sigma$) of the number of views taken to cover 80\% of the reconstruction for each type of shoe. The first three shoes are from the training and the last three are from the test set.}
\label{tab:robot_experiment}
\end{table}


\section{Conclusion}
In this paper we presented a novel method for view planning to cover the visible surface of an object and perform multi-view 3D reconstruction.
Our method leverages recent learning-based reconstruction techniques to guide planning for the camera locations and incorporate multiple measurements.
Specifically, our method estimates an intermediate global geometry of the object and uses this information to choose the next viewpoints.
As the camera obtains more measurements, the predicted shape of the object is refined and the object's visible surface is covered.

Our experimental analysis revealed that the reconstruction uncertainty is related to visibility and using intermediate reconstructions provide valuable information for view planning.

Possible future directions include testing our method on a more comprehensive dataset containing larger number of different categories and shapes, and investigating the advantage of more efficient view planning in the context of a full mobile manipulation system that incorporates the estimated 3D reconstructions with grasp and manipulation planning and control.

\bibliography{references}
\bibliographystyle{unsrt}

\end{document}